\documentclass{article}

\usepackage[preprint]{neurips_2026}

\usepackage[utf8]{inputenc}
\usepackage[T1]{fontenc}
\usepackage{amsfonts}
\usepackage{amsmath, amssymb}
\usepackage{txfonts}
\usepackage[hidelinks]{hyperref}
\usepackage{url}
\usepackage{booktabs}
\usepackage{nicefrac}
\usepackage{microtype}
\usepackage{xcolor}
\usepackage{multirow}
\usepackage{graphicx}
\graphicspath{{./}}

\title{Position: Mechanistic Interpretability Must Disclose Identification Assumptions for Causal Claims}

\author{%
Zezheng Lin\thanks{Correspondence to Zezheng Lin.} \And Fengming Liu }

\begin{document}
\maketitle

\begin{abstract}
Mechanistic interpretability papers increasingly use causal vocabulary: circuits, mediators, causal abstraction, monosemanticity. Such claims require explicit identification assumptions. A purposive audit of 10 papers across four methodological strands finds no dedicated identification-assumptions section and a recurring pattern: validation metrics such as faithfulness, completeness, monosemanticity, alignment, or ablation effects are reported as causal support without stating the assumptions that make them identifying. A two-human-coder audit on $n=30$ reproduces the direction of the main finding: dedicated identification sections are absent, and validation-metric substitution is common, though exact Dim B/D counts are coding-rule sensitive. The paper proposes a disclosure norm: state whether the claim is causal, name the identification strategy, enumerate assumptions, stress at least one, and explain how conclusions shift if assumptions fail. Validation is not identification.
\end{abstract}

\section{Introduction}
\label{sec:intro}

\textbf{Position.} \emph{Mechanistic interpretability research using causal language must disclose identification assumptions. A paper making a causal claim must state in its abstract whether the claim is causal, name the identification strategy, enumerate the assumptions under which the claim is identified, test or stress at least one core assumption, and state how conclusions shift if key assumptions fail.}

The mechanistic interpretability literature has already produced its own evidence that this matters. \citet{wang2023iwild} presented a prominent ICLR 2023 paper identifying a GPT-2 circuit for indirect object identification using activation patching. The paper did not state the assumptions under which the circuit claim was identified. A year later, \citet{makelov2024illusion} diagnosed one such failure mode and showed that naive subspace patching can produce interpretability illusions: the patched direction may not be the direction the behavior actually depends on. Their paper is careful and self-aware; what it does not do is generalize the diagnosis into a field-wide assumption-disclosure protocol that other papers can apply before failure shows up downstream. \citet{leask2025sae} produced an analogous correction for sparse autoencoders, showing that approximately one-third of features in a larger SAE are absent from a smaller one trained on the same data---direct empirical pressure on the dictionary-basis-recoverability assumption that monosemanticity work \citep{bricken2023monosemanticity, templeton2024scaling} relied on without stating. \citet{canby2024probing} formalized completeness and selectivity for causal probing as competing desiderata, surfacing assumptions that the underlying probing literature had not declared.

This sequence is not unique to a few cases. Causal vocabulary (circuits, mediators, causal abstraction, monosemanticity) is adopted enthusiastically; the assumptions under which the causal claim is identified are left implicit; downstream work later diagnoses a specific identification failure; and the field absorbs the correction as a methodological hazard rather than as evidence that disclosure should have come first. The audit reported here documents this across four major methodological strands.

\paragraph{Validation is not identification.} The repeated move in the audit is what
this paper calls \emph{validation metric substitution}: faithfulness, completeness, monosemanticity, behavioral alignment, or alignment-search accuracy are reported in support of causal claims without a separate statement of the identification assumptions those claims require. A faithful, complete, monosemantic, or behaviorally aligned description is consistent with the model and with itself, but consistency does not establish that the proposed mechanism is the cause rather than a correlate. \citet{makelov2024illusion} and \citet{canby2024probing}, both in the sample, demonstrate empirically that high validation scores can co-exist with failed identification.

\begin{figure}[h]
\centering
\includegraphics[width=0.88\linewidth]{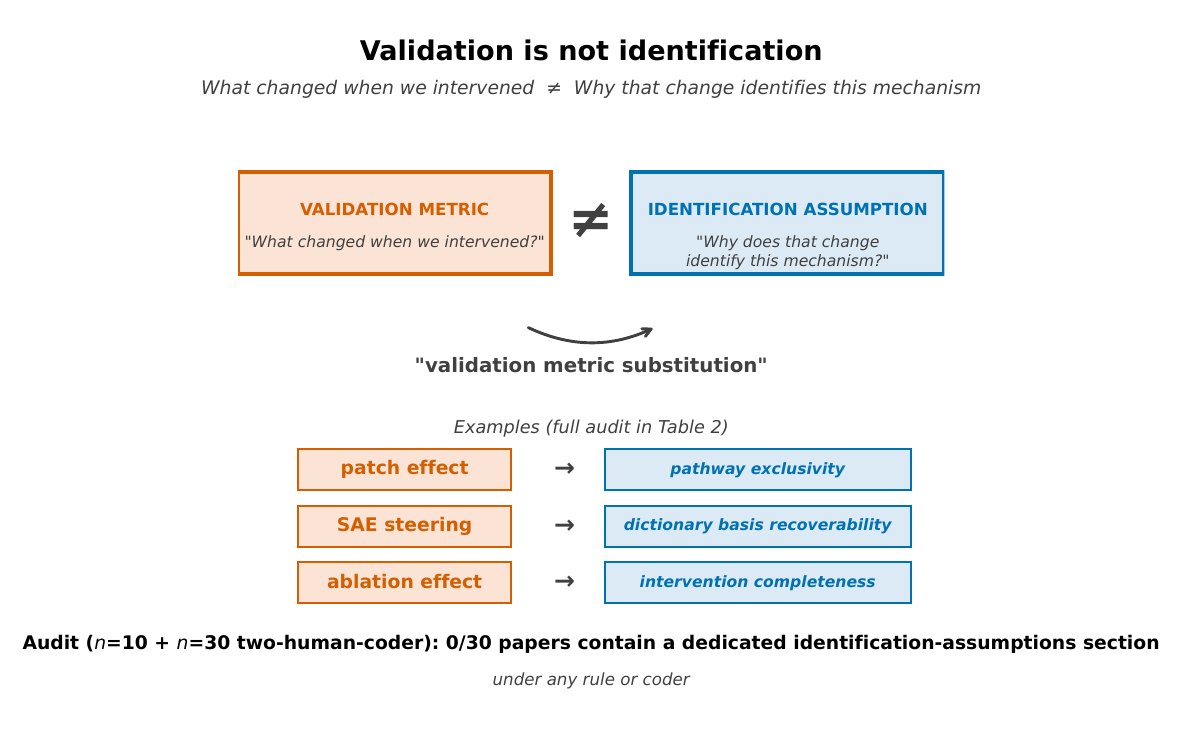}
\caption{Validation metrics and identification assumptions are not interchangeable.
Validation metrics report what the data show under assumed conditions;
identification assumptions are the conditions under which a causal claim follows.
Substitution---reporting the metric in place of the assumption---leaves the
causal claim unidentified. Audit ($n=10$ + $n=30$ sensitivity): $0/30$ papers
contain a dedicated identification-assumptions section under any rule or coder.}
\label{fig:val_vs_id}
\end{figure}

\paragraph{Why borrow a disclosure norm from causal inference.} Econometrics did not
eliminate identification disputes; it institutionalized the norm that causal claims are evaluated through stated assumptions \citep{haavelmo1944, heckman1979, angrist1996iv, imbens2020potential}. The discipline this paper proposes does not import instrumental variables, regression discontinuity, or any specific causal framework into MI. It imports the meta-practice: when a paper makes a causal claim, it states the assumptions under which the claim is identified, and it offers a falsifiability test or sensitivity statement for at least one of them. This is a weaker requirement than committing to potential outcomes, structural causal models, or do-calculus, and it is generative rather than restrictive: explicit assumptions enable cumulative critique.

\paragraph{Audit and templates.} The audit covers papers spanning four strands across
NeurIPS, ICLR, CLeaR, BlackboxNLP, and the Anthropic Transformer Circuits Thread. The aggregate pattern is that no paper contains a dedicated identification-assumptions section, the majority make only implicit causal claims in abstracts, the majority contain no falsifiability test, and the majority substitute a validation metric for an identification statement. The paper then offers method-specific templates that authors can copy into their Methods sections for activation patching, sparse autoencoders, causal abstraction, and probing with ablation, listing assumptions, the evidence that addresses them, potential falsifiers, and the sensitivity implications if violated.

\paragraph{Scope of the audit.} The audited sample is purposive and small, and is
meant to support the position rather than to estimate prevalence in the broader MI literature. Per-paper evidence (abstract summaries, source locators, main-text search evidence, justification text) is released so that every coding can be challenged on the published rubric. The position does not depend on prevalence statistics; it depends on the conceptual point that disclosure of identification assumptions is a logical requirement of any causal claim. The audit functions as evidence that the gap is widespread enough across major venues and strands to warrant a community standard.

\section{Related Work}

Several recent papers have analyzed evaluation practices in machine learning, though none have focused specifically on identification assumption disclosure in mechanistic interpretability. \citet{schaeffer2023emergent} showed that claimed emergent abilities often vanish under careful evaluation, demonstrating that the ML community's evaluation practices can produce conclusions that fail under scrutiny. \citet{pearl2009causality, pearl2018book} established the formal foundations of causal inference and argued that causal claims require explicit assumptions; this work provides the theoretical backbone for our claim that LLM interpretability papers should state their identification assumptions.

More closely related, \citet{leask2025sae} challenged sparse-autoencoder feature atomicity empirically, showing that features decompose across SAE sizes in ways that invalidate claims about individual feature causality. They trained sparse autoencoders of different sizes on the same model and inserted latents from a larger SAE into a smaller one, finding that approximately one-third of features in the larger SAE were absent from the smaller one, suggesting larger SAEs capture information the smaller ones miss. This finding directly challenges the dictionary-basis-recoverability assumption: if no canonical basis exists, the SAE decomposition is arbitrary up to rotation, and attributing causal behavior to individual features is unjustified. This is the only prior work that directly identifies an unstated identification assumption in interpretability and tests it. Our contribution is to make this kind of assumption-checking the standard practice rather than a post-hoc discovery.

\section{The Econometrics Precedent}

In econometrics, a paper claiming a causal effect of treatment $D$ on outcome $Y$ states in its opening section the identification strategy supporting the claim. This is not a stylistic choice but a methodological one: without stated assumptions, the scientific community cannot evaluate whether a result reflects genuine causal structure or an artifact of unstated conditions.

In the potential outcome framework \citep{rubin1974potential, holland1986statistics}, a causal claim requires specifying the stable unit treatment value assumption (SUTVA), which states that the outcome for one unit is unaffected by the treatment assignment of other units, and the treatment assignment mechanism, typically unconfoundedness (ignorability), which states that potential outcomes are independent of treatment assignment conditional on observed covariates. In the DAG framework \citep{pearl2009causality}, causal identification requires specifying the causal graph and the Markov assumptions, including d-separation criteria.

The two frameworks differ in technical detail but converge on the same meta practice: explicit assumption disclosure is a prerequisite for any causal claim, regardless of framework. This is not merely a disciplinary convention. It is a logical requirement of causal inference: a causal claim has no meaning without specifying the conditions under which the causal quantity is identified from observable data.

For LLM interpretability, the parallel is direct in the meta sense. Activation patching assumes interventions isolate causal pathways. Sparse autoencoders assume learned features correspond to the model's true representational basis. Causal abstraction assumes distributed representations align with higher-level causal structures. Probing assumes the representational space admits localized causal interventions. These are identification assumptions: conditions that must hold for causal claims to follow. Yet the LLM interpretability literature almost never states them.

\section{Identification Assumptions in Mechanistic Interpretability}
\label{sec:assumptions}

This section enumerates the most common identification assumptions per methodological strand in the audited sample. The point is not to prescribe a fixed list but to show that such assumptions exist for every method and can be made explicit.

\paragraph{Activation patching.} (1) \emph{Circuit completeness}: the targeted components
account for all causal pathways. (2) \emph{Pathway exclusivity}: patching does not activate dormant parallel circuits. (3) \emph{Metric sufficiency}: the behavioral metric reflects the causal property of interest. \citet{wang2023iwild} patched 10 attention heads in GPT-2 Small for indirect object identification; they did not state that the discovered circuit exhausts all causal pathways. \citet{makelov2024illusion} patched subspaces in various models and assumed the patched subspace was the sole driver of the effect. They did not enumerate this assumption.

\paragraph{Sparse autoencoders.} (1) \emph{Dictionary basis recoverability}: the SAE
decomposes activations into the model's true feature basis rather than an arbitrary rotation. (2) \emph{Feature atomicity}: features are non-decomposable into simpler sub-features. (3) \emph{Monosemanticity}: each feature corresponds to a single coherent concept rather than a polysemantic blend. (4) \emph{Feature completeness}: the SAE captures all causally relevant features. \citet{bricken2023monosemanticity} attributed behavior to individual features in GPT-2; they assumed those features were atomic and monosemantic. \citet{templeton2024scaling} scaled this approach to Claude 3 Sonnet, assuming the larger model's feature space decomposes into a canonical basis. None of these assumptions were formally stated. The consequences are visible in subsequent work: \citet{leask2025sae} showed that dictionary basis recoverability fails empirically.

\paragraph{Causal abstraction.} (1) \emph{Alignment faithfulness}: the structural
correspondence between neural representations and higher-level causal models is preserved under intervention. (2) \emph{Optimization validity}: the gradient-based search procedure finds genuine alignments rather than statistical artifacts. \citet{wu_geiger2023das} applied distributed alignment search to Alpaca-7B and found alignments with two-boolean causal models. \citet{geiger2024clear} formalized alignment using conditional independence and morphism compatibility. Neither assumption was declared as a numbered statement.

\paragraph{Probing.} (1) \emph{Causal locality}: the relevant representation is concentrated
in individual neurons or small subspaces rather than distributively encoded. (2) \emph{Intervention completeness}: ablation removes the full representational content of the target concept. (3) \emph{Behavioral sufficiency}: downstream behavioral effects reflect the conceptual content being tested. \citet{huang2023neurons} ablated single GPT-2 XL neurons; they assumed single neurons were appropriate causal units. Both assumptions were implicit. \citet{canby2024probing} formalized completeness and selectivity as competing desiderata---the most explicit framing in our sample---but stopped short of formally declaring identification assumptions.

\section{Audit Methodology}
\label{sec:audit_method}

\subsection{Sample Selection}

The audit covers ten papers spanning four methodological strands (circuit discovery, sparse autoencoders, causal abstraction, probing) and the venues NeurIPS, ICLR, CLeaR, BlackboxNLP, and the Anthropic Transformer Circuits Thread. The sample includes pioneering works (\citealt{conmy2023acdc}, \citealt{bricken2023monosemanticity}) and recent advances (\citealt{templeton2024scaling}, \citealt{dunefsky2024transcoders}). The selection is purposive rather than random; findings are not prevalence estimates. The consistency across strands and years is read as evidence that the pattern is not restricted to outliers, with the caveat that purposive sampling cannot establish prevalence in the broader literature.

\subsection{Coding Protocol}

Each paper is coded on four dimensions:
\begin{itemize}
\item \textbf{Dimension A (Abstract claim type)}: I = explicit causal w/ ID basis;
II = explicit causal w/o ID basis; III = implicit causal; IV = association only.
\item \textbf{Dimension B (Main-text identification assumptions)}: I = dedicated
Identification Assumptions section; II = partial discussion; III = no explicit statement.
\item \textbf{Dimension C (Falsifiability)}: Yes = falsifiability test; Partial = sensitivity
analysis only; No = none.
\item \textbf{Dimension D (Validation-metric substitution)}: Yes = a validation metric
(faithfulness, completeness, monosemanticity, alignment, etc.) is treated as causal support
without a separate identification-assumption statement; Partial = the validation metric is
discussed with caveats but its relation to identification is not explicitly separated;
No = the paper clearly distinguishes validation criteria from identification assumptions.
\end{itemize}

The full coding rubric, including decision rules for ambiguous cases and anchor examples, is in the released audit package (\texttt{coding\_decision\_rules.md}). The audit is \textbf{transparent and single-coder}: for each paper the coder records a short abstract summary, source locator, main-text search evidence, and a justification paragraph, then assigns codes on the four dimensions. Cohen's $\kappa$ and other inter-rater statistics are not reported for the $n=10$ audit because there is only one coder at this stage; computing $\kappa$ from a single coder would be unsound. Two-rater human validation is reported separately for the extension audits below. The released rubric and per-paper evidence support independent re-coding of any individual entry.

\subsection{Decision Rule for the ``Causal Method Label vs.\ Causal Claim'' Distinction}

The hardest call in coding Dim A is whether use of ``causal interventions'' or ``causal model'' constitutes a causal claim (II) or a method label (III). The uniform rule:

\begin{table}[h]
\centering
\caption{Decision rule for Dim A coding.}
\small
\begin{tabular}{p{0.55\linewidth}c}
\toprule
\textbf{Case} & \textbf{Code} \\
\midrule
``Causal interventions'' describes patching as a method, no causal estimand or identification basis & III \\
Explicit ``we causally identify X'' or ``causal effect of X on Y'', no ID strategy named & II \\
Explicit causal claim + named ID strategy + assumptions in abstract & I \\
No causal-adjacent language at all & IV \\
\bottomrule
\end{tabular}
\end{table}

Under a more permissive rule---where any use of the word ``causal'' promotes a paper to II---three currently III-coded papers (\citealt{wu_geiger2023das}, \citealt{templeton2024scaling}, \citealt{bricken2023monosemanticity}) could be re-coded II. The change does not affect the main argument: II papers also fail the disclosure protocol because they make causal claims without disclosing identification assumptions, and no paper in the sample reaches I under either rule.

\section{Audit Results}
\label{sec:audit_results}

\begin{table}[h]
\centering
\caption{Audit results: ten mechanistic interpretability papers. Coding column abbreviations
match Section~\ref{sec:audit_method}. Per-paper evidence (short abstract summaries,
source locators, main-text search evidence, and per-paper justifications) appears in the
released audit package.}
\label{tab:audit}
\small
\begin{tabular}{llcccc}
\toprule
\textbf{Paper} & \textbf{Strand} & \textbf{Dim A} & \textbf{Dim B} & \textbf{Dim C} & \textbf{Dim D} \\
\midrule
\citet{conmy2023acdc}            & Circuit discovery   & III & III & No      & Yes     \\
\citet{wang2023iwild}            & Circuit discovery   & III & II  & No      & Yes     \\
\citet{wu_geiger2023das}         & Causal abstraction  & III & II  & No      & Partial \\
\citet{geiger2024clear}          & Causal abstraction  & II  & II  & Partial & Partial \\
\citet{templeton2024scaling}     & Sparse autoencoders & III & III & No      & Yes     \\
\citet{bricken2023monosemanticity} & Sparse autoencoders & III & III & No    & Yes     \\
\citet{makelov2024illusion}      & Activation patching & III & II  & Partial & No      \\
\citet{huang2023neurons}         & Probing             & III & III & No      & Partial \\
\citet{dunefsky2024transcoders}  & Sparse autoencoders & III & II  & No      & Yes     \\
\citet{canby2024probing}         & Probing             & II  & II  & Partial & No      \\
\bottomrule
\end{tabular}
\end{table}

\paragraph{Aggregate findings.} Across the sample:
\begin{itemize}
\item Dim A: 0/10 explicit causal with ID basis; 2/10 explicit without; 8/10 implicit; 0/10 association-only.
\item Dim B: 0/10 dedicated Identification Assumptions section; 6/10 partial discussion; 4/10 no explicit statement.
\item Dim C: 0/10 falsifiability test; 3/10 partial sensitivity; 7/10 no falsifiability evidence.
\item Dim D: 5/10 treat a validation metric as causal support without separating it from
identification (Yes); 3/10 discuss the validation metric with caveats but do not separate
it from identification (Partial); 2/10 clearly separate validation from identification
(No, both papers whose primary contribution is methodological critique).
\end{itemize}

The Dim D distribution is concentrated: 8 of 10 papers in the sample either fully or partially substitute a validation metric (faithfulness, completeness, monosemanticity, alignment search success) for an identification statement. The two papers that cleanly separate the two are \citet{makelov2024illusion} and \citet{canby2024probing}, both of whose primary contribution is to demonstrate failure modes of the validation criteria. This concentration is consistent with the reading that within the sample, identification-validation separation is not a baseline practice but arises only when separation itself is the research question.

These counts are consistent with a community-wide pattern of causal language without identification disclosure across the sampled methodological strands and venues, while the small purposive sample limits prevalence claims for the broader literature.

\paragraph{Anchor case: \citet{makelov2024illusion}.} The most methodologically
self-aware paper in the sample. Patching what appeared to be the correct direction for a known feature produced the expected behavioral change; activating an orthogonal direction in the same subspace produced the same change, showing the effect was not uniquely attributable to the feature direction. Even this paper does not formally enumerate the assumptions underlying its critique. The implicit assumptions are the standard activation-patching set: the target subspace captures all causally relevant contributions; patching does not activate parallel pathways; the behavior is locally rather than distributively implemented. \citet{makelov2024illusion} come closest in the sample to testing identification assumptions empirically; the assumptions are described rather than declared.

\paragraph{Sparse autoencoder consequences.} \citet{leask2025sae} ran stitching
experiments showing that approximately a third of features in larger SAEs are absent from smaller ones, directly challenging dictionary basis recoverability; meta-SAE experiments showed that larger-SAE features decompose into combinations of smaller-SAE features, challenging feature atomicity. Both findings invalidate assumptions that the audited SAE papers never stated, so they could not anticipate or hedge against these failure modes.

\section{Two-Human-Coder Audit ($n=30$)}
\label{sec:two_coder}

To probe robustness of the $n=10$ single-coder audit and estimate inter-coder agreement, a two-human-coder audit was conducted on $n=30$ MI papers spanning the four strands plus circuit-discovery scaling work, drawn from NeurIPS, ICLR, ICML, EMNLP, NAACL, CLeaR, BlackboxNLP, NeurIPS workshops, the Anthropic Transformer Circuits Thread, and arXiv. Per-paper records, raw responses from both coders, and the $\kappa$/$\alpha$ scripts are released in the audit package.

\paragraph{Coding protocol.} Two independent bilingual coders (Coder A
and Coder B) coded the $n=30$ records using the same rubric and identical paper metadata (title, strand, venue, year, abstract). The coders did not see each other's labels and did not see the $n=10$ main-audit codings. $\kappa$ and $\alpha$ in Table~\ref{tab:irr} measure human inter-rater reliability between the two coders. The $n=30$ extension is coded from title, abstract, strand, venue, and year. This is appropriate for auditing whether identification disclosure appears in the most visible reporting layer that readers and reviewers see under standard reading practice; it is not a full-text recoding of all assumptions, and the headline counts in Section~\ref{sec:two_coder} should be read on that basis.

\begin{table}[h]
\centering
\small
\caption{Two-human-coder agreement on $n=30$ MI papers (human
inter-rater reliability). $P_o$: raw agreement; $\kappa$: Cohen's $\kappa$
between two human coders using the same rubric;
$\alpha$: Krippendorff's $\alpha$ (nominal, two-coder).}
\label{tab:irr}
\begin{tabular}{lccc}
\toprule
Dimension & $P_o$ & $\kappa$ & $\alpha$ \\
\midrule
Dim A (abstract claim type)        & $0.80$ & $0.56$ & $0.57$ \\
Dim B (assumption disclosure)      & $0.57$ & $0.26$ & $0.14$ \\
Dim C (falsifiability)             & $0.77$ & $0.52$ & $0.51$ \\
Dim D (validation substitution)    & $0.77$ & $0.43$ & $0.40$ \\
\bottomrule
\end{tabular}
\end{table}

\begin{figure}[h]
\centering
\includegraphics[width=0.92\linewidth]{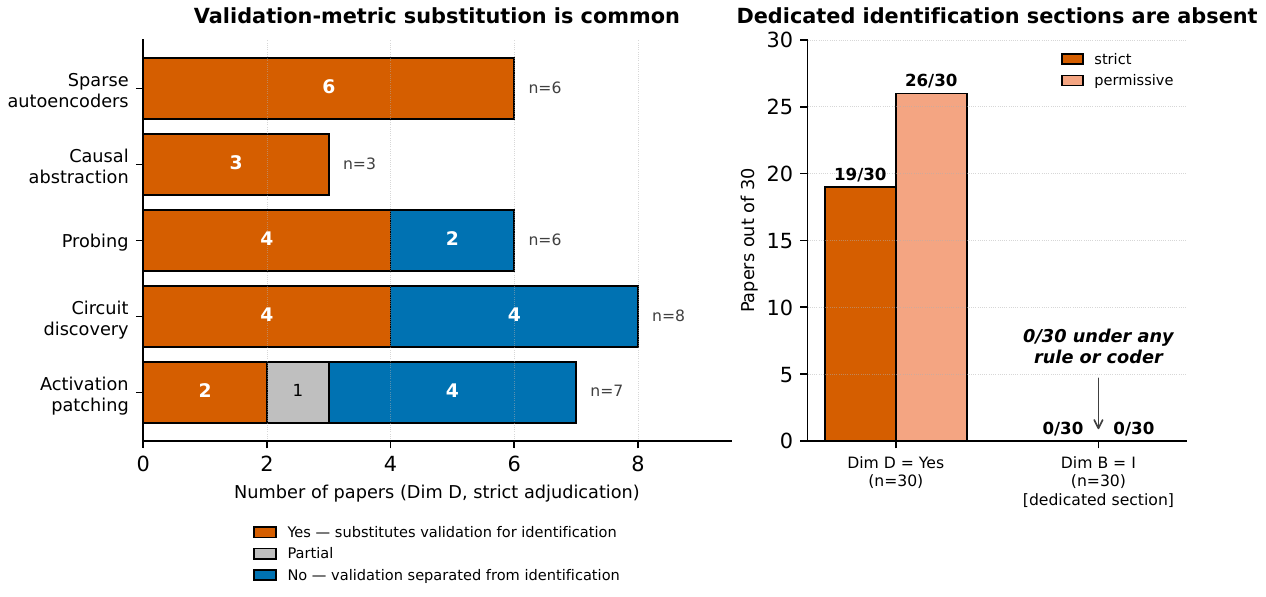}
\caption{$n=30$ two-human-coder audit (human inter-rater
reliability). \emph{Left}: Dim D under strict adjudication, by methodological
strand. Sparse autoencoders ($6/6$) and causal abstraction ($3/3$) substitute
validation for identification at full rate; activation patching shows the
lowest rate ($2/7$), reflecting partial protective effect of the
illusion-and-failure-modes literature within that strand. \emph{Right}:
Dim D = Yes occurs on $19/30$ papers under the strict rule and $26/30$ under
the permissive rule; Dim B (dedicated identification-assumptions section) is
$0/30$ under any rule or coder. Counts are from the abstract-layer audit;
exact Dim D counts are coding-rule sensitive.}
\label{fig:dimD_killer}
\end{figure}

\paragraph{Findings.} Agreement is moderate on Dim A and C and lower on
Dim B and Dim D. The low Dim B agreement is itself substantive: coders disagreed on whether informal limitation language counts as partial assumption disclosure ($P$ vs $N$). This ambiguity strengthens rather than weakens the disclosure recommendation---current papers do not provide stable, easily identifiable assumption statements, which is exactly what the position is asking the field to fix. On Dim B, no paper is coded $I$ under either rule or coder. Disagreements on Dim D were resolved twice: a strict rule (less-substitution wins) yields Dim D $= \text{Yes}$ on $19/30$ papers, a permissive rule yields $26/30$. The Dim D count should be read as a moderate-confidence structural pattern, not a precise prevalence estimate. Per-strand strict counts and the strict-vs-permissive sensitivity are reported in Figure~\ref{fig:dimD_killer}.

\paragraph{What this changes about the $n=10$ finding.} The $n=30$ audit
reproduces the direction of all $n=10$ findings ($0/30$ dedicated-assumption sections, $18$--$22/30$ implicit-causal abstracts, $19$--$26/30$ Dim D under strict-vs-permissive sensitivity) and surfaces strand heterogeneity the $n=10$ obscured: SAE and causal-abstraction strands show stronger substitution than activation patching, where the illusion-and-failure-modes literature has a partial protective effect within that strand.

\paragraph{Author-conducted human audit on $n=20$ (Dim B + Dim D).} A
single-rater bilingual human audit of Dim B and Dim D was performed by the paper author on a stratified sub-sample of $n=20$ from the $n=30$ paper list (5 Circuit discovery, 5 Activation patching, 4 Sparse autoencoders, 4 Probing, 2 Causal abstraction; seed \texttt{0xB1D}). Source material was abstracts only, matching the $n=30$ audit's information access. The author was blind to the $n=30$ codings at coding time. Headline counts: Dim B $= I$ (dedicated section): $\mathbf{0/20}$; Dim D $= \text{Yes}$ (substitution): $\mathbf{16/20}$. Author-vs-strict-adjudication agreement: Dim B $16/20$ ($80\%$), Dim D $13/20$ ($65\%$). The headline direction of both findings is reproduced by the single human rater; the $0/30$ dedicated-section result is reproduced exactly. Raw codes and per-paper rationales are released in \texttt{audit\_package\_paper2/human\_audit\_n20\_codes.json}.

\paragraph{Identification limits and second human-coder sensitivity.} The
human-coder audit ($n=30$) and single-rater human audit ($n=20$) agree on the structural finding ($0/30$ and $0/20$ dedicated identification-assumption sections) and on the direction of the substitution count. An additional second human-coder re-coding of Dim B/D is released as human two-coder validation. On the $n=20$ first-coder sample the second coder gives agreement $80\%$ ($\kappa=0.49$) on Dim B and $70\%$ ($\kappa=0.46$) on Dim D, with the second coder more conservative (some Dim D labels move from Yes to Partial). On the $n=30$ records the second coder differs more from the existing strict adjudication ($\kappa=0.08$ on Dim B, $\kappa=0.25$ on Dim D), with the second coder using Partial much more often. The $n=30$ audit and the second-coder re-coding read abstracts and strand metadata only; full-text re-coding by independent external coders is not included in this submission. The two-human-coder $\kappa$ on Dim B ($0.26$) is low enough that the precise $19$--$26/30$ Dim D count is better read as a moderate-confidence point estimate than as a prevalence measurement; the $0/30$ structural finding is not threatened by this and is also reproduced by the human raters.

\section{The Disclosure Protocol}
\label{sec:protocol}

The protocol adapts the causal-inference convention of opening-section assumption disclosure to mechanistic interpretability. It does not prescribe a causal framework; it requires that whatever framework a paper adopts, its core assumptions be stated explicitly. The recommendation to NeurIPS, ICLR, ICML, and comparable venues is to incorporate the protocol into existing submission checklists. The five requirements:

\begin{enumerate}
\item State in the abstract whether the paper makes a causal claim or an association claim.
Vague language that could be read either way is not acceptable.
\item If a causal claim appears in the abstract, name the identification strategy supporting it.
\item Include a numbered Identification Assumptions section in the main text. Each assumption
must state what it asserts, why the authors consider it plausible, and what evidence exists
for it.
\item For each core assumption, provide at least one of: a direct falsifiability test within
the experimental setup, a sensitivity analysis showing how conclusions change under violation,
or a clear statement of what empirical finding would count against it.
\item Discuss substantively how conclusions change if key assumptions fail.
\end{enumerate}

\paragraph{Example application to \citet{wang2023iwild}.} The abstract would
state a causal claim supported by activation patching under circuit completeness and pathway exclusivity. The main text would enumerate: the discovered circuit exhausts indirect-object-identification pathways; patching does not activate parallel circuits; the logit-difference metric captures the causal essence of the behavior. Assumptions 1 and 2 would be tested. \citet{makelov2024illusion} raised exactly the parallel-pathway concern, yet \citet{wang2023iwild} never engaged with it; the protocol gives reviewers a common ground for the question.

\paragraph{Validation metrics versus identification assumptions.} Mechanistic
interpretability papers commonly report validation metrics that establish empirical relevance but do not test the assumptions needed for the causal claim. Table~\ref{tab:val_id} summarizes four common method-metric pairs and the identification gap each one leaves open.

\begin{table}[h]
\centering
\caption{Common validation metrics in mechanistic interpretability and the identification
assumptions they do not test. Reporting these metrics is necessary but not sufficient for
the causal claims the methods are typically used to support.}
\label{tab:val_id}
\small
\begin{tabular}{p{2.5cm}p{3.0cm}p{3.0cm}p{3.5cm}}
\toprule
Method & Validation metric typically reported & Identification assumption left untested & Why the metric does not test it \\
\midrule
Activation patching & patch effect / logit difference & pathway exclusivity, circuit completeness & a measurable patch effect can flow through dormant or parallel pathways \\
SAE steering & behavioral change under feature activation & canonical basis, feature atomicity & steering shows a feature is causally relevant, not that the basis is canonical or features atomic \\
Causal abstraction (DAS / IIA) & alignment score & optimization validity, alignment faithfulness & high IIA can reflect spurious alignments learned by the optimizer rather than a faithful abstraction \\
Probing with ablation & ablation effect & causal locality, intervention completeness & a null effect can reflect distributed representation rather than absence of the underlying structure \\
\bottomrule
\end{tabular}
\end{table}

\section{Limitations and Alternative Views}

\paragraph{Limitations.} The $n=10$ audit is purposive and single-coder; it
establishes a documented gap across major strands, not a prevalence estimate. The $n=30$ extension uses two human coders, and the $n=20$ Dim B/D check uses one human rater; the released second human-coder re-coding of Dim B/D provides human two-coder validation on that subset. Because all extension codings use abstracts and metadata rather than full texts, exact Dim B/D counts should be read as coding-rule-sensitive; the robust structural finding is the absence of dedicated identification-assumptions sections and the recurrence of validation-metric substitution.

\paragraph{Alternative views.} One view is that MI is engineering rather than
causal science. The protocol applies only when papers use causal vocabulary; non-causal papers can avoid the requirement by framing claims descriptively. A second view is that formal assumption disclosure is premature. The response is that the protocol does not impose econometric estimators or a single causal framework; it asks authors to state the assumptions their own interventions already require. A third view is that boilerplate sections could become performative. Reviewer guidance should therefore distinguish substantive assumptions, falsifiers, and sensitivity statements from perfunctory checklists. The supplement provides method-specific templates to make the norm concrete.

\bibliographystyle{plainnat}

\end{document}